\documentclass[letterpaper]{article} 
\usepackage[preprint]{aaai2027}  
\usepackage[hyphens]{url}  
\usepackage{graphicx} 
\usepackage{natbib}  
\usepackage{caption} 
\usepackage{algorithm}
\usepackage{algorithmic}

\usepackage{newfloat}
\usepackage{listings}
\DeclareCaptionStyle{ruled}{labelfont=normalfont,labelsep=colon,strut=off} 
\floatstyle{ruled}
\newfloat{listing}{tb}{lst}{}
\floatname{listing}{Listing}

\usepackage{booktabs}

\usepackage{amsmath}

\title{Zero-Fi: Zero-Shot Wi-Fi-Based Human Activity Recognition via Contrastive Signal-Language Alignment}

\author {
    Yitong Shen\textsuperscript{\rm 1}\equalcontrib,
    Cheng Guo\textsuperscript{\rm 2}\equalcontrib,
    Peiliang Wang\textsuperscript{\rm 1},
    Jingzhe Zhang\textsuperscript{\rm 1},
    Yi Sheng\textsuperscript{\rm 1},
    Haopeng Zhang\textsuperscript{\rm 2},
    Hongfei Xue\textsuperscript{\rm 2}\corresponding,
    Yili Ren\textsuperscript{\rm 1}\corresponding
}
\affiliations {
    \textsuperscript{\rm 1}University of South Florida\\
    \textsuperscript{\rm 2}University of North Carolina at Charlotte\\
    shen202@usf.edu, cguo3@charlotte.edu, peiliangwang@usf.edu, jingzhe@usf.edu, sheng1@usf.edu, haopeng.zhang@charlotte.edu, hongfei.xue@charlotte.edu, yiliren@usf.edu
}

\newcommand{\methodname}{Zero-Fi}

\begin{document}

\maketitle

\begin{abstract}

Wi-Fi-based human activity recognition has advanced substantially, but most existing methods assume a closed set of activities and require labeled Wi-Fi samples for every target class, limiting their ability to recognize unseen activities. We present Zero-Fi, a contrastive signal-language alignment framework for zero-shot Wi-Fi-based human activity recognition. Zero-Fi learns unified representations from complementary Wi-Fi signal features and aligns them with the semantic representations of natural-language activity descriptions in a shared embedding space. This cross-modal alignment enables Zero-Fi to recognize new activity classes without requiring labeled Wi-Fi samples or model adaptation for those classes. Experiments on large-scale public benchmark datasets demonstrate effective zero-shot recognition of held-out activity classes, highlighting the potential of signal-language alignment to extend Wi-Fi sensing beyond predefined activity classes.
\end{abstract}




\section{Introduction}

Human activity recognition (HAR) is an important technology that automatically detects and identifies human physical movements~\cite{kaur2024human}. Over the decades, HAR has been extensively studied~\cite{zhang2020human} and has demonstrated broad applicability in various areas such as healthcare~\cite{ge2022contactless}, security~\cite{jiang2013rejecting}, human-computer interaction~\cite{kellogg2014bringing}, and smart homes~\cite{du2019novel}.


Owing to its substantial practical value, HAR has been realized through various conventional sensing modalities, including cameras~\cite{tran2018closer}, inertial measurement units (IMUs)~\cite{zhang2022if}, wearable sensors~\cite{ordonez2016deep}, millimeter-wave (mmWave) radar~\cite{singh2019radhar}, and LiDAR~\cite{luo2020temporal}. However, each modality has inherent limitations. 
Specifically, IMU- and wearable-based systems require users to continuously carry or wear dedicated devices/sensors, which can be intrusive and inconvenient. Radar- and LiDAR-based systems~\cite{cui2021high, bouazizi20232d} often incur non-negligible hardware costs and deployment complexity. The performance of camera-based HAR systems~\cite{zhang2019comprehensive, liu2019deep} is highly sensitive to illumination conditions and may deteriorate substantially in low-light or dark environments.



In recent years, Wi-Fi has emerged as a promising sensing modality for HAR. Human movements alter the propagation paths of Wi-Fi signals through reflection, diffraction, and scattering, producing measurable variations in signals~\cite{tan2022commodity}. These variations encode motion-related information that can be analyzed to distinguish different human activities. Compared with conventional sensing modalities, Wi-Fi enables contactless sensing and remains robust under varying ambient illumination conditions. Moreover, existing Wi-Fi infrastructure and widely available Wi-Fi devices can be repurposed for HAR, thereby reducing deployment costs. The pervasive coverage of Wi-Fi signals in indoor environments further provides a practical foundation for scalable and widely deployable HAR systems.

Owing to these advantages, Wi-Fi-based HAR has attracted growing research interest, leading to the development of numerous systems~\cite{adib2013see, wang2015understanding, zheng2019zero, xiao2021onefi, li2021two, zhang2026wi}. However, most existing systems have limited ability to generalize to previously unseen activity categories.
They typically rely on annotated training data and strong supervision under closed-set assumptions, and thus cannot recognize activities that are absent from the training set.
This limitation reduces their practical applicability because deployed HAR systems may encounter activity categories that are absent from the training data.
Given the breadth and diversity of human behavior, collecting and annotating representative Wi-Fi samples for every possible activity category is prohibitively expensive and ultimately infeasible. The ability to recognize unseen activities is therefore essential for Wi-Fi-based HAR systems. 


In this paper, we present \methodname{}, a novel framework for zero-shot Wi-Fi-based human activity recognition that can effectively recognize unseen activities.
Realizing this capability, however, requires addressing three key challenges.

First, zero-shot Wi-Fi-based HAR requires learning representations from Wi-Fi signals that generalize beyond the activity categories observed during training. The key challenge is to learn Wi-Fi representations that capture shared, transferable motion characteristics, rather than features specific to a fixed set of predefined activities.
To address this challenge, we align Wi-Fi representations with the semantic space of natural language using contrastive learning, which offers a scalable interface for describing human activities~\cite{guadarrama2013youtube2text}, including categories absent from the Wi-Fi training data. Unlike discrete labels that treat each activity as an independent category, language descriptions explicitly encode motion attributes, such as body parts, movement directions, and speeds, that are shared across activities. Although Wi-Fi signals and language differ substantially in form, both characterize the same underlying human motion: Wi-Fi signals capture the physical effects of body movements on wireless signal propagation, while language expresses those movements at a semantic level. Aligning the two modalities therefore enables the model to associate Wi-Fi signals with these shared motion attributes.
Since unseen activities can often be characterized as new combinations of attributes already observed in seen activities, the resulting signal-language correspondence naturally extends beyond the training classes, enabling zero-shot recognition.

Second, Wi-Fi signals are inherently sensitive to environmental variations. Since the signal component reflected off the human body is often superimposed with signals reflected from surrounding objects, even minor environmental changes (e.g., furniture relocation) can cause the received signals to differ substantially for the same activity. In addition, Wi-Fi devices introduce hardware-induced phase noise, such as random phase offset and carrier frequency offset~\cite{ma2019wifi}, which further distorts the signal. The challenge is therefore to reliably extract activity-relevant features from Wi-Fi signals that remain robust to such environmental and hardware noise.
To address these, we introduce a domain discriminator that encourages the feature encoder to suppress environment-specific information while preserving motion patterns. To suppress hardware-induced noise, we leverage the fact that distortions are shared across antennas on the same Wi-Fi device, and cancel them out by computing the conjugate multiplication of signals received across antennas, thereby removing the common noise component while preserving activity-relevant signal variations.

Third, existing Wi-Fi HAR datasets typically provide only simple activity labels. Directly aligning Wi-Fi representations with such brief text labels is unlikely to yield satisfactory generalization, as simple labels fail to capture detailed motion attributes and the subtle relationships among activities. For example, ``answering the phone'' and ``combing hair'' are semantically distant as class names, yet both activities involve moving the arm toward the head. The challenge is therefore to obtain semantically rich language descriptions that explicitly characterize the underlying human motion.
To address this, we leverage the broad semantic knowledge encoded in large language models (LLMs) to decompose each activity label into motion attributes associated with the torso, head, arms, hands, legs, and overall trajectory. This allows us to explicitly model the semantic relationships between activities that share common motion components: although ``answering the phone'' and ``combing hair'' are distinct as class names, their attribute-level descriptions both involve moving the arm toward the head, resulting in high similarity between their attribute embeddings.

We evaluate \methodname{} on public datasets under a strict zero-shot setting, in which the Wi-Fi signals and language descriptions of all test activity categories are unseen during training. Results demonstrate that \methodname{} effectively recognizes unseen activities, achieving an average accuracy of 69.58\%, and consistently outperforms existing baselines.
Our main contributions are summarized as follows:
\begin{itemize}
\item We present \methodname{}, a contrastive framework for zero-shot Wi-Fi-based human activity recognition that aligns Wi-Fi signals with semantically rich language descriptions.
\item We design mechanisms to extract robust Wi-Fi features and to construct rich, attribute-level language descriptions for each activity, enabling cross-modal alignment. 
\item We conduct extensive experiments on multiple public datasets, demonstrating the effectiveness and generalization of our work under strict zero-shot settings.
\end{itemize}

\section{Related Work}

\subsection{Human Activity Recognition}



HAR has been implemented using diverse sensing modalities~\cite{kaur2024human}. For example, IF-ConvTransformer~\cite{zhang2022if} fuses measurements from multiple IMU sensors and employs a convolutional Transformer for activity recognition. TSAM~\cite{li2025frame} augments a pretrained visual backbone with a sequential perceiver adapter to capture spatial and temporal features. mmCLIP~\cite{cao2024mmclip} aligns mmWave radar heatmaps with textual embeddings to recognize unseen activities, while Luo et al.~\cite{luo2020temporal} combine LSTM and temporal convolutional networks to classify activities from LiDAR point clusters. However, these modalities face inherent limitations in deployment cost, hardware requirements, and robustness to illumination conditions. In contrast, Wi-Fi-based HAR offers a practical and scalable alternative by leveraging the ubiquitous Wi-Fi devices and signals.

\subsection{Wi-Fi Sensing}

Wi-Fi sensing has been widely studied for object sensing~\cite{ren2020liquid, wang2024wi2dmeasure}, localization~\cite{kotaru2015spotfi, fan2024multitarget}, pose estimation~\cite{ren2022gopose, yan2024person}, security~\cite{jiang2013rejecting, lin2023contactless}, healthcare~\cite{ge2022contactless, wang2024target}, smart homes~\cite{du2019novel, li2025wilife}, and HAR~\cite{li2021two, xiao2021onefi, li2025consense, zhang2026wi}. For HAR, THAT~\cite{li2021two} models time-channel dependencies using a two-stream convolution-augmented Transformer, while OneFi~\cite{xiao2021onefi} applies meta-learning~\cite{hospedales2021meta} for one-shot gesture adaptation. However, they cannot support strict zero-shot recognition. Wi-Chat~\cite{ren2025wichat} uses an LLM to infer activities without task-specific training, but its reliance on manually designed signal-description prompts limits its demonstrated zero-shot capability to four activity categories.

\section{Preliminary}

\subsection{Wi-Fi Sensing Basics}

\begin{figure}[!t]
    \centering
    \includegraphics[width=0.68\columnwidth]{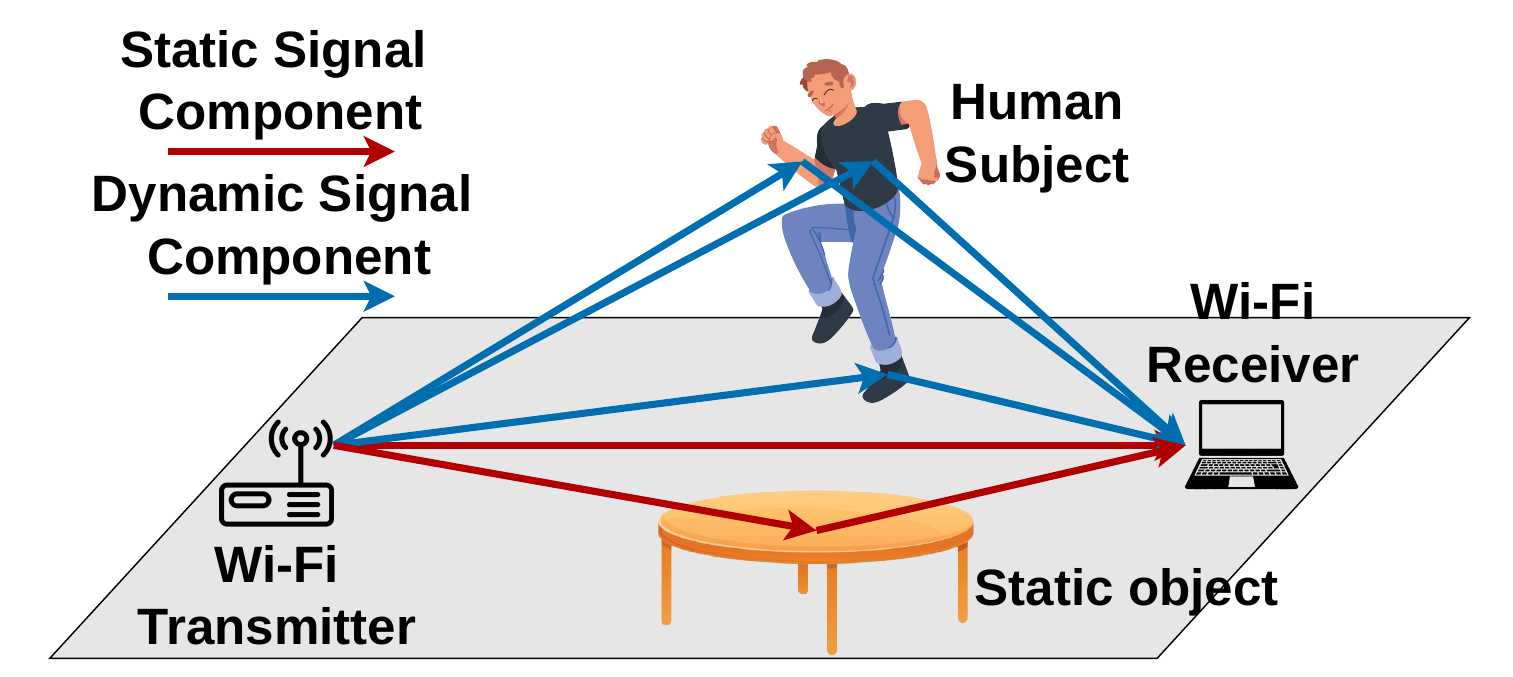}
    \caption{Wi-Fi sensing illustration. 
    }
    \label{sensing}
\end{figure}

Wi-Fi has evolved beyond its traditional role as a communication technology to become a promising sensing modality~\cite{tan2022commodity}.
As illustrated in Figure~\ref{sensing}, a Wi-Fi transmitter emits signals that propagate through the environment, interact with the human body through reflection, diffraction, and scattering, and are subsequently captured by a Wi-Fi receiver. Human movements alter these propagation paths, producing measurable variations in the received signals. Because different activities generate distinct signal patterns, these variations can be analyzed to infer human motion and recognize activities.

\subsection{Wi-Fi Channel State Information}

To capture Wi-Fi signal changes caused by a target, we utilize the Channel State Information (CSI). 
It describes how a transmitted signal propagates through the wireless medium and captures the effects of human movements and the environment. 
Wi-Fi CSI could be expressed as $H(f, t) = \sum_{i}^{N}A_{i}e^{-j2\pi\frac{d_{i}(t)}{\lambda}}.$
Here, $A_{i}$ is the complex attenuation, $d_{i}(t)$ is the length of the $i^{th}$ path, $N$ is the total number of paths, and $\lambda$ is the wavelength.
As shown in Figure~\ref{sensing}, the received Wi-Fi signal can be divided into static and dynamic components. Specifically, static components are the signal reflections from static objects (e.g., furniture) and dynamic components are the signals reflected off the dynamic human body.
Thus, CSI can be further expressed as
$
    H(f, t) = H_{s}(f, t) + H_{dyn}(f, t) 
            = H_{s}(f, t) + a(f, t)e^{-j2\pi\frac{d(t)}{\lambda}}
$,
where $H_{s}(f,t)$ and $H_{dyn}(f,t)$ denote the static and dynamic components, respectively. $a(f,t)$ represents the amplitude attenuation of the dynamic component, $e^{-j2\pi\frac{d(t)}{\lambda}}$ denotes the corresponding phase, $d(t)$ is the propagation path length of the dynamic component, and $\lambda$ is the signal wavelength.

\section{Methodology}
\label{Methodology}

\begin{figure*}[!t]
    \centering
    \includegraphics[width=0.9\textwidth]{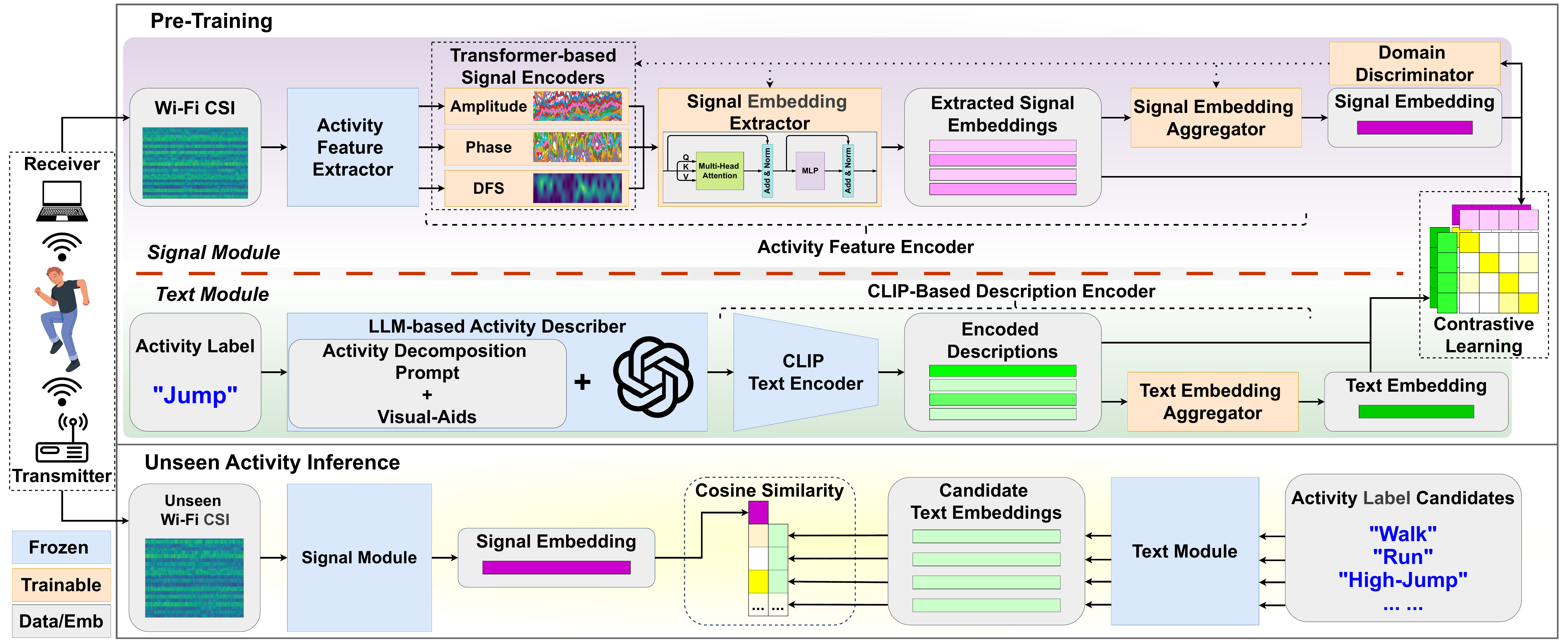}
    \caption{Overview of \methodname{}.}
    \label{Overview}
\end{figure*}



As shown in Figure~\ref{Overview}, \methodname{} comprises signal, text, and signal-text contrastive learning modules.


\subsection{Signal Module}
\label{SignalModule}

\textbf{Activity Feature Extractor.}
Commodity Wi-Fi devices introduce hardware noise including random phase offset (RPO) and carrier frequency offset (CFO)~\cite{kotaru2015spotfi, li2016dynamic}.
In particular, RPO and CFO induce time-varying phase offsets that distort the temporal phase variations of CSI, thereby degrading the fidelity of the extracted signal representations.
Accordingly, Wi-Fi CSI can be modeled as
$
    H(f, t) = e^{-j\theta}(H_{s}(f, t) + a(f, t)e^{-j2\pi\frac{d(t)}{\lambda}}),
$
where $e^{-j\theta}$ represents the aggregation of these phase offsets.
Because these phase offsets are the same across the antennas of the same Wi-Fi device, conjugate multiplication of CSIs between antennas can be applied to suppress them:
\begin{align}
    H_{cm}(f, t) & = H_{1}(f, t)\Bar{H}_{2}(f, t) \notag \\
                 & = (e^{-j\theta}(H_{1, s}(f, t) + a_{1}(f, t)e^{-j2\pi\frac{d_{1}(t)}{\lambda}})) \notag \\ & (e^{j\theta}(\Bar{H}_{2, s}(f, t) + a_{2}(f, t)e^{j2\pi\frac{d_{2}(t)}{\lambda}})) \notag \\
                 & = H_{1, s}(f, t)\Bar{H}_{2, s}(f, t) \quad \cdots \quad \textcircled{1} \notag \\
                 & + a_{1}(f, t)a_{2}(f, t)e^{-j2\pi\frac{d_{1}(t) - d_{2}(t)}{\lambda}}  \quad \cdots \quad \textcircled{2} \notag \\
                 & + \Bar{H}_{2, s}(f, t)a_{1}(f, t)e^{-j2\pi\frac{d_{1}(t)}{\lambda}}))   \quad \cdots \quad \textcircled{3} \notag \\
                 & + H_{1, s}(f, t)a_{2}(f, t)e^{j2\pi\frac{d_{2}(t)}{\lambda}}. \quad \cdots \quad \textcircled{4} 
\end{align}
Here, $H_{cm}(f, t)$ denotes the result of conjugate multiplication, while $H_{1}(f, t)$ and $\Bar{H}_{2}(f, t)$ denote the CSI obtained from one antenna and the complex-conjugated CSI obtained from another antenna, respectively. 
Term $\textcircled{1}$ is motion-invariant and removed by high-pass filtering. Term $\textcircled{2}$ has a small magnitude and is therefore negligible, and terms $\textcircled{3}$ and $\textcircled{4}$ retain the dominant motion-induced variations.
The denoised CSIs are subsequently used for feature extraction.

\begin{figure}[!t]
    \centering



    \includegraphics[width=0.32\columnwidth]{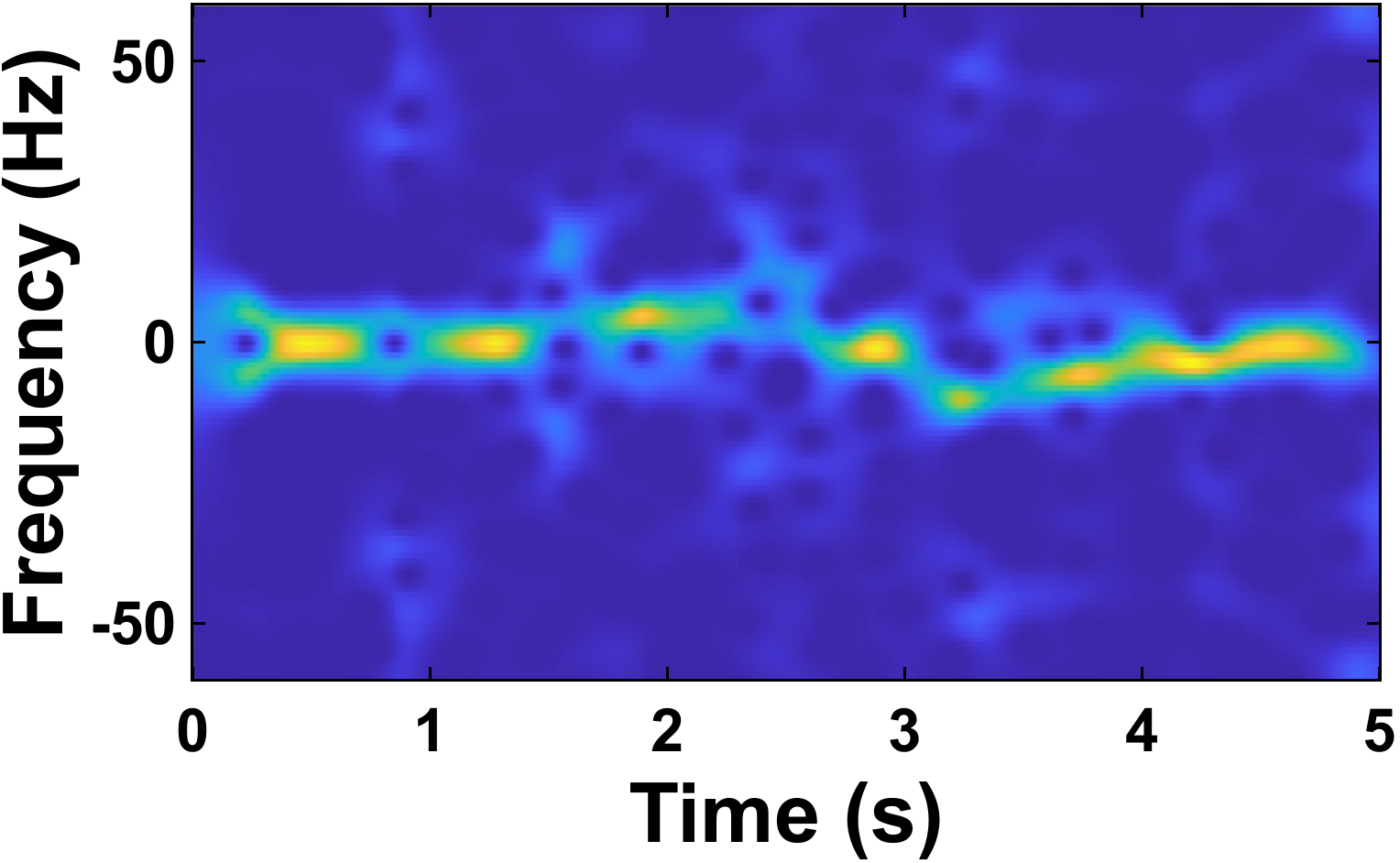}
    \includegraphics[width=0.32\columnwidth]{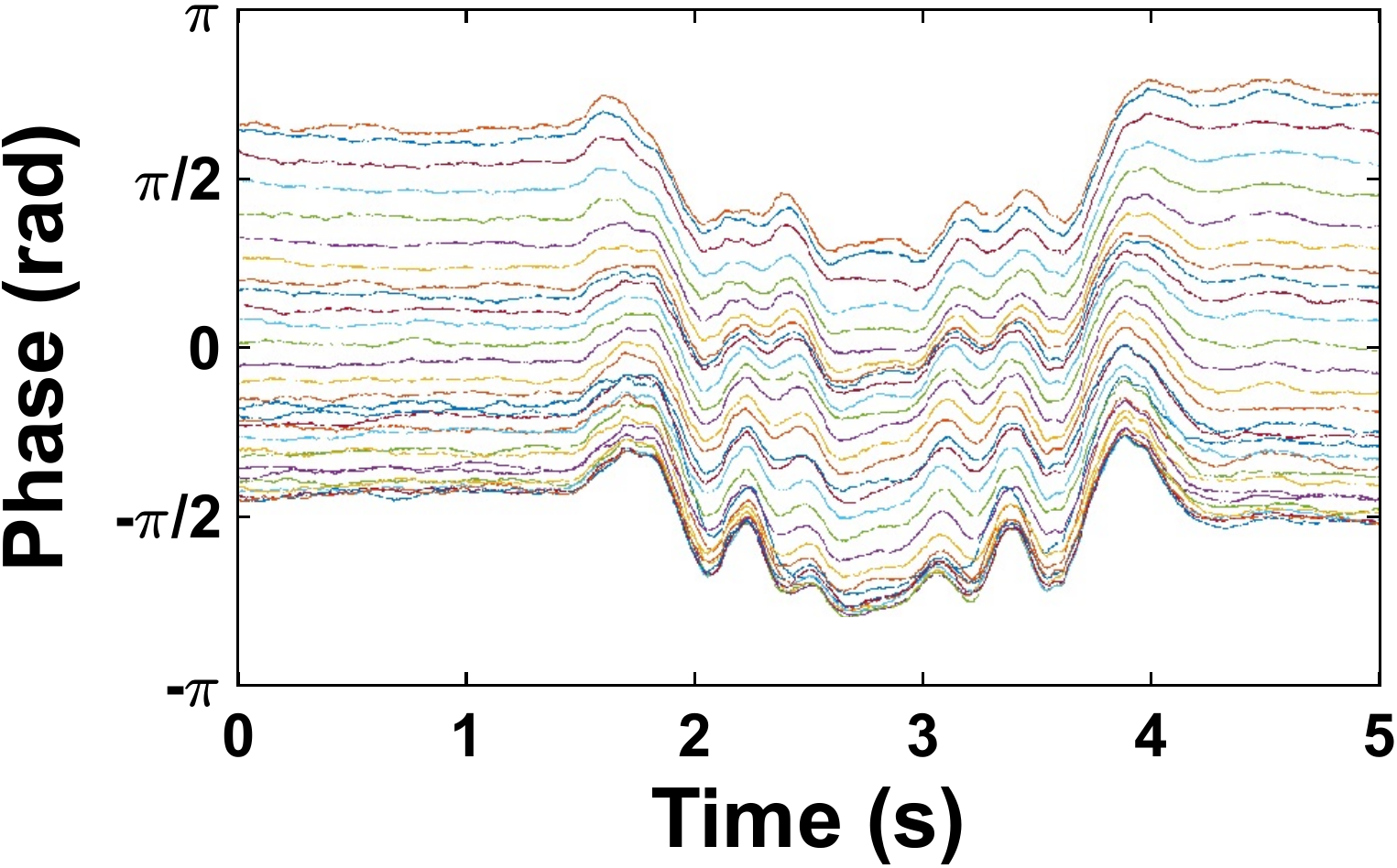}
    \includegraphics[width=0.32\columnwidth]{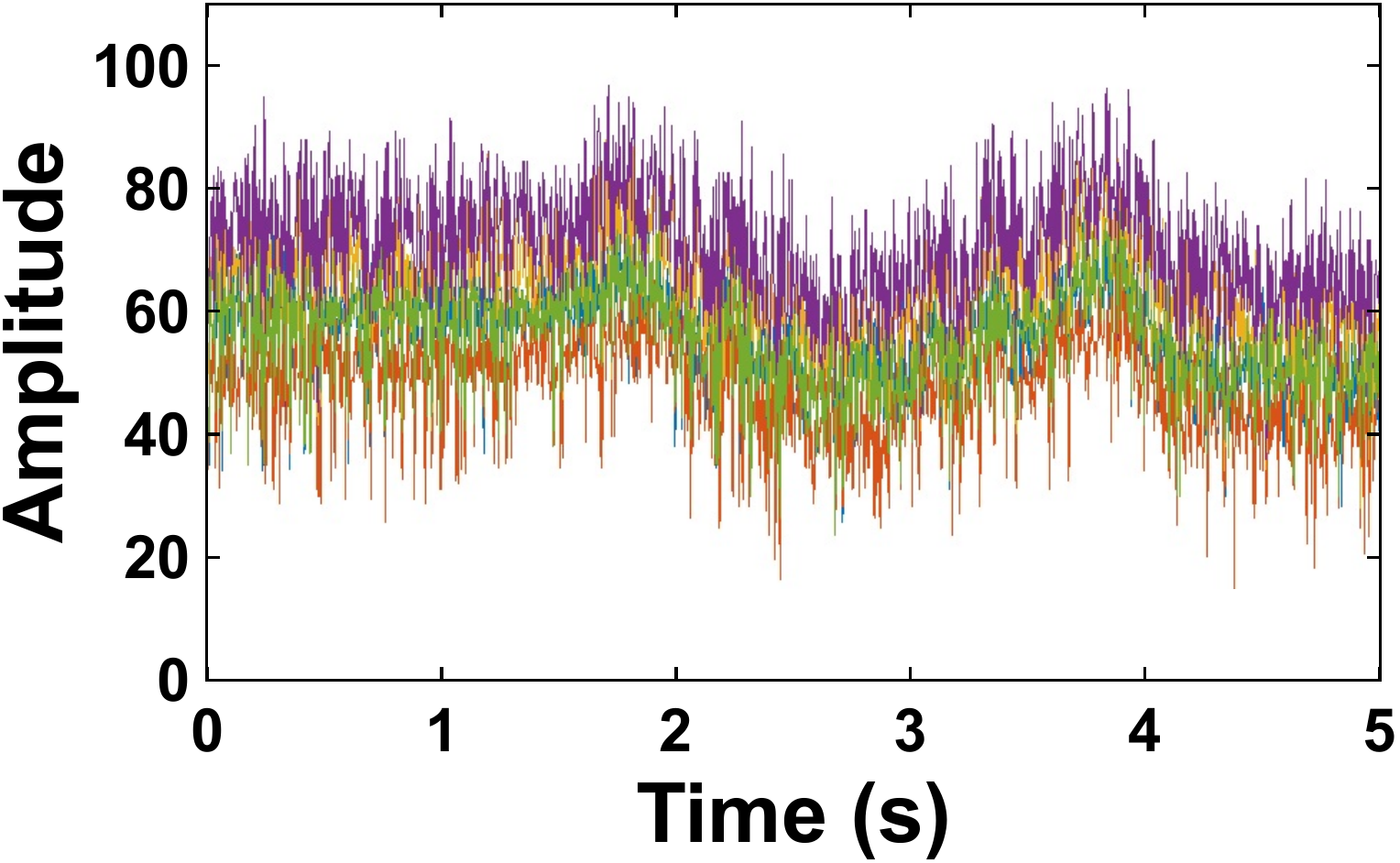}
    \caption{DFS, phase, and amplitude features.
    }
    \label{Extracted}
\end{figure}

We then extract three complementary representations from Wi-Fi signals: Doppler frequency shift (DFS), phase, and amplitude. DFS captures the time-frequency dynamics of human motion, phase preserves fine-grained activity variations, and amplitude reflects coarse-grained motion patterns~\cite{qian2017widance, ren2020liquid}.
For DFS extraction,
we first compute the maximum mean-to-variance ratio of the amplitude for each transmit-receive antenna pair's CSI.
The CSIs with higher amplitude values and lower amplitude variance tend to contain less dynamic information~\cite{qian2017widance}. 
Therefore, removing these CSIs helps retain signals with richer motion-related dynamics. 
The Short-Time Fourier Transform (STFT) is subsequently applied to obtain the time-frequency representation, from which the frequency bins within the target DFS range are
retained~\cite{zheng2019zero}.
An additional $\ell_{1}$ normalization step is applied to the extracted DFS representations to reduce scale differences across samples.
The signal phase can be extracted using the $angle(\cdot)$ function as follows
$
    H_{P} = unwrap(angle(H_{cm}(f, t))).
$
Because phase measurements are periodic with a period of $2\pi$, the $unwrap(\cdot)$ function is applied to eliminate discontinuities caused by phase wrapping. 
Then, a sliding-window smoothing filter is applied to $H_{P}$ to reduce residual fluctuations and improve the stability of the extracted phase.
The signal amplitude $H_{A}$ can be calculated as $H_{A} = abs(H(f, t)).$
We also apply a sliding-window smoothing filter to $H_{A}$ to improve the stability of the amplitude.
The extracted Wi-Fi representations for human activities are shown in Figure~\ref{Extracted}.

\textbf{Domain Discriminator.} 
CSIs are affected by both dataset-specific acquisition conditions and environment-specific characteristics. 
Because these factors are often coupled in public Wi-Fi sensing datasets, we treat each unique dataset-environment configuration as an independent domain. 
To mitigate the influence of configuration-specific characteristics, we introduce a domain discriminator that encourages the activity feature encoder in Figure~\ref{Overview} to learn domain-invariant activity representations.
Each training sample $x_{i}$ is therefore associated with a domain label $d_{i} \in \{1,\ldots, N_{d}\}$, where $N_{d}$ denotes the number of domains represented in the training data. 
Let $\mathbf{z}_{i} = F_{AFE}(x_i)$ denote the signal embedding produced by the activity feature encoder $F_{AFE}(\cdot)$. 
We first calculate its cosine similarity $sim(\cdot)$ to the text embedding $\mathbf{t}_c$ of each seen activity class:
$
    \mathbf{q}_i = \operatorname{softmax}(\exp(\tau) \operatorname{sim} \left(\bar{\mathbf{z}}_i, \bar{\mathbf{t}}_c\right)),
$
where $\bar{\mathbf{z}}_i$ and $\bar{\mathbf{t}}_c$ denote $\ell_2$-normalized signal and text embeddings, respectively, $\tau$ is the learnable log-temperature parameter, and $\mathbf{q}_i$ is the predicted activity distribution over the seen classes. Only descriptions of the seen activities are used to calculate $\mathbf{q}_i$ during training.

The domain discriminator is conditioned on both the signal representation and its predicted activity distribution~\cite{zhao2017ei}:
$
\hat{\mathbf{p}}_i^{\,d}
=
D_{\phi}(\operatorname{GRL}_{\gamma_d}(\mathbf{z}_i)
\oplus
\operatorname{sg}(\mathbf{q}_i)),
$
where $\operatorname{GRL}_{\gamma_d}(\cdot)$ denotes a gradient-reversal layer with coefficient $\gamma_d$, $\operatorname{sg}(\cdot)$ denotes the stop-gradient operation, $\oplus$ denotes concatenation, and $D_{\phi}$ denotes the domain discriminator. Conditioning the discriminator on the activity prediction helps account for activity-dependent differences between domains, while stopping the gradient through $\mathbf{q}_i$ prevents the domain objective from directly altering the activity predictions.
The discriminator is trained using the domain-classification loss
\begin{equation}
    \mathcal{L}_{\mathrm{domain}}
    =
    -\frac{1}{B}
    \sum_{i=1}^{B}
    \sum_{j=1}^{N_d}
    \mathcal{I}[d_i=j]
    \log \hat{p}_{i,j}^{\,d},
\end{equation}
where $B$ denotes the batch size, $\mathcal{I}[\cdot]$ is the indicator
function, and $\hat{p}_{i,j}^{\,d}$ is the predicted probability that sample
$i$ belongs to dataset-environment configuration $j$. We use domain-balanced mini-batches to reduce the influence of domain imbalance.
The overall training objective is
\begin{equation}
    \mathcal{L}_{\mathrm{total}} = \mathcal{L}_{\mathrm{con}} + \mathcal{L}_{\mathrm{domain}}.
\end{equation}
During backpropagation, the gradient-reversal layer multiplies the gradient from $\mathcal{L}_{\mathrm{domain}}$ to the activity feature encoder by $-\gamma_d$. 
Thus, the domain discriminator is optimized to correctly identify the dataset-environment configuration, whereas the activity feature encoder is optimized to make this prediction difficult while preserving signal-text alignment.
This adversarial objective encourages the learned signal representations to become less dependent on configuration-specific acquisition and propagation characteristics.

\textbf{Activity Feature Encoder.}
We design three distinct, representation-specific Transformer-based signal encoders, one for each extracted CSI representation: amplitude, phase, and DFS.
For the amplitude and phase encoders, the two-dimensional convolutional layer commonly adopted in standard Vision Transformer (ViT) architectures~\cite{dosovitskiy2020image} is replaced with a one-dimensional convolutional layer. 
This design captures the temporal structure of the amplitude and phase representations while preserving dependencies among their constituent measurements, which contain discriminative activity-related information.
Specifically, given the amplitude and phase inputs $H_{A}, H_{P} \in \mathcal{R}^{t \times N_{sc}}$, where $t$ denotes the temporal length and $N_{sc}$ denotes the number of subcarriers, the corresponding encoders apply one-dimensional convolutional patch embedding, temporal positional encoding, and self-attention to each representation. 
The resulting embeddings for the encoded amplitude and phase representations are denoted by $Emb_{A}, Emb_{P} \in \mathcal{R}^{\lfloor t/s \rfloor \times d_{SE}}$, respectively, where $s$ denotes the temporal downsampling factor and $d_{SE}$ denotes the output feature dimension of the signal encoders.
For the DFS encoder, we retain the standard ViT architecture because the STFT-derived DFS jointly captures both temporal and frequency-domain features.
For the DFS input $H_{D} \in \mathcal{R}^{t \times N_{f} \times 1}$, where $N_{f}$ denotes the number of frequency bins, a two-dimensional convolutional patch embedding with a window size of $w \times h$ is applied. The result is projected to an embedding $Emb_{D} \in \mathcal{R}^{\lfloor t / w \rfloor \times d_{SE}}$.

The encoded amplitude, phase, and DFS embeddings have identical dimensions, $\mathcal{R}^{T \times d_{SE}}$, where $T=\lfloor t/w \rfloor$ denotes the temporal length.
All three representations preserve temporal information, and the amplitude and phase are inherently sequential; their encoded embeddings are concatenated along the feature dimension while maintaining temporal correspondence across representations.
A learnable positional encoding $PE(\cdot)$ is subsequently applied along the temporal dimension to capture temporal dependencies:
$Emb = PE(Emb_{D} \oplus Emb_{P} \oplus Emb_{A})$.
Here, $Emb$ denotes the concatenated embedding, and $\oplus$ denotes concatenation along the feature dimension.
Consequently, the resulting embedding $Emb$ has dimensions $(T, 3d_{SE})$.
Next, a signal embedding extractor (SEE) is used to learn shared representations from the concatenated modality embeddings.
The SEE employs multi-head self-attention to model cross-modal interactions before forwarding the resulting features to the subsequent aggregation stage.
The attention output is further processed using layer normalization $LN(\cdot)$ and a multilayer perceptron $MLP(\cdot)$, together with a residual connection: $z = Attention(Q, K, V),$ $Emb_{SEE} = MLP(LN(z)) + z.$
Here, the query $Q$, key $K$, and value $V$ are obtained through separate linear projections of the concatenated embedding $Emb$.
Finally, we employ a single transformer layer, termed the signal embedding aggregator (SEA), to further aggregate and refine the representations produced by the SEE:
$
    Emb_{signal} = Linear(SEA(Emb_{SEE})).
$
Here, $Linear(\cdot)$ denotes a linear projection layer. The resulting representation $Emb_{\mathrm{signal}}$ serves as the final CSI-derived embedding.

\subsection{Text Module}



\begin{figure}[!ht]
    \centering
    \includegraphics[width=0.9\columnwidth]{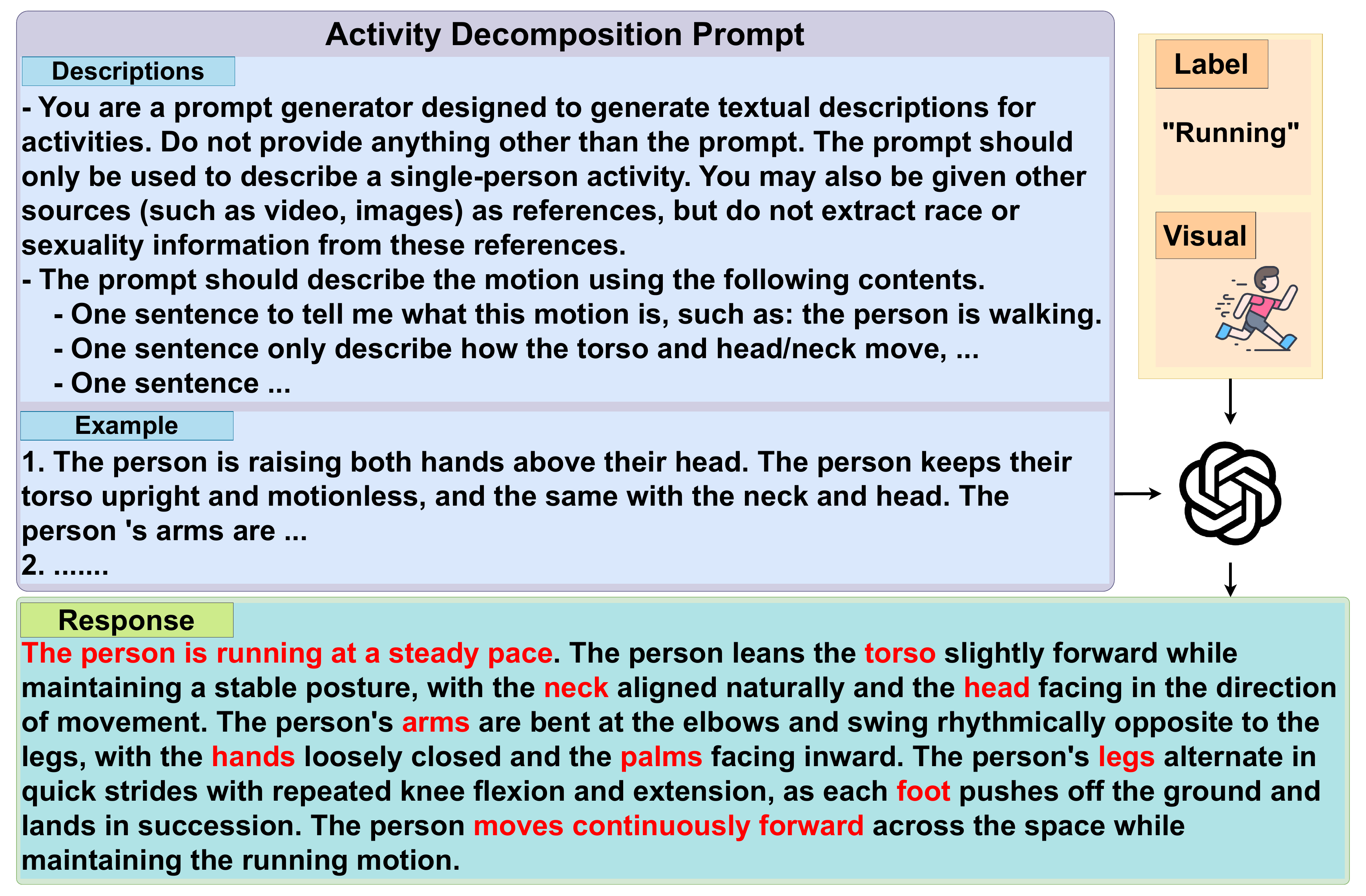}
    \caption{LLM-based activity describer.}
    \label{ActivityDescriber}
\end{figure}

\textbf{LLM-Based Activity Describer.}
Effective signal-language alignment requires richer semantic supervision than that provided by a single activity label. 
We therefore employ an LLM, pretrained on large-scale text corpora, to construct the activity describer.
The architecture of this component is illustrated in Figure~\ref{ActivityDescriber}.
To constrain and standardize its output, we design a structured prompt that instructs the LLM to expand each activity label into a detailed description of body-part movements, including those of the head, arms, hands, torso, and legs, as well as the overall movement direction and trajectory. 
Specifically, the activity label and the structured prompt are jointly provided to the LLM to generate the corresponding activity description. To further improve descriptive accuracy, we exploit the multimodal reasoning capabilities of contemporary LLMs by supplying relevant visual references, such as images and videos, when available.
These descriptions provide semantically informative representations of both seen and unseen daily activities, thereby facilitating subsequent alignment with the corresponding signal features.

\textbf{CLIP-Based Description Encoder.}
Because language descriptions and Wi-Fi CSI measurements are not inherently aligned, their representations must be projected into a shared latent space. 
CLIP~\cite{radford2021clip} was originally trained to align textual and visual representations through contrastive learning. 
Its text encoder therefore captures visually grounded semantic information related to objects, actions, and human motion, which can provide a useful supervisory representation for activity-related signal features. 
Accordingly, we employ the frozen CLIP text encoder to encode the generated activity descriptions for subsequent contrastive alignment with the CSI representations. The resulting text embeddings are denoted by $Emb_{t-enc}$.
In addition, we introduce a single transformer layer, termed the Text Embedding Aggregator (TEA), to aggregate and refine the embeddings of multiple descriptions into a unified representation, denoted by $Emb_{text}$, for downstream tasks.

\subsection{Signal-Text Alignment}

The objective of this stage is to train the overall framework to produce semantically meaningful signal embeddings and align them with the corresponding textual representations.
This alignment is achieved through contrastive learning, which increases the similarity between positive signal-text pairs associated with the same activity while decreasing the similarity between negative pairs associated with different activities.
Specifically, we employ the InfoNCE objective~\cite{radford2021clip} to encourage bidirectional alignment between the signal and textual embeddings. 
The signal embedding$Emb_{signal}$ and text embedding $Emb_{text}$ are first $\ell_2$-normalized.
Their pairwise similarities are computed as
$
    \ell_{i,j}
    =
    \exp(\tau)\,
    \operatorname{sim}
    \left(
        \frac{\mathrm{Emb}_{\mathrm{signal}}^{i}}
        {\|\mathrm{Emb}_{\mathrm{signal}}^{i}\|_2},
        \frac{\mathrm{Emb}_{\mathrm{text}}^{j}}
        {\|\mathrm{Emb}_{\mathrm{text}}^{j}\|_2}
    \right),
$
where $\tau$ is a learnable log-temperature parameter, $sim(\cdot,\cdot)$ denotes cosine similarity, and $\ell_{i,j}$ denotes the similarity between the $i$th signal sample and the $j$th text sample. The resulting similarity matrix has dimensions $B\times B$.
To reduce the occurrence of false-negative pairs caused by multiple samples from the same activity class, we employ class-aware sampling to maximize activity-class diversity within each batch.
Consequently, most batches contain at most one sample from each activity class, although duplicate classes may occasionally occur because of sampling constraints.

Cross-entropy losses are then computed in both alignment directions: text-to-signal and signal-to-text. The overall contrastive loss is defined as
\begin{equation}
    L_{con} = \frac{1}{2}(a L_{CE, t2s} + b L_{CE, s2t}),
\end{equation}
where $L_{CE, t2s}$ and $L_{CE, s2t}$ denote the text-to-signal and signal-to-text contrastive losses, respectively. The weighting coefficients $a$ and $b$ are set to $1$ by default.
Each directional loss is implemented using cross-entropy:
$
    L_{CE}
    =
    -\frac{1}{B}
    \sum_{i=1}^{B}
    \sum_{j=1}^{B}
    y_{i,j}\log \hat{y}_{i,j},
$
where $B$ denotes the batch size, $y_{i,j}=1$ when the $i$th signal and $j$th text form the paired sample, i.e., $i=j$, and $y_{i,j}=0$ otherwise. 
The predicted probability $\hat{y}_{i,j}$ is obtained by applying softmax along the corresponding signal-to-text or text-to-signal direction of the similarity matrix.
The contrastive loss is applied not only to the final aggregated embeddings but also to the intermediate signal and text embeddings, $Emb_{SEE}$ and $Emb_{t-enc}$, respectively. 
This intermediate-level supervision enables the signal encoders to learn more directly from the textual embeddings. 
Through bidirectional contrastive supervision at both the intermediate and final embeddings, the model receives semantic guidance from both the less-processed features and the refined embeddings.

\section{Experiment}


\subsection{Datasets}


We evaluate our method on two of the largest Wi-Fi datasets, WiDAR 3.0~\cite{zheng2019zero} and XRF55~\cite{wang2024xrf55}, which are combined for training and zero-shot testing.
\textbf{WiDAR 3.0} is one of the largest publicly available datasets for human activity recognition. 
It contains approximately 270,000 samples covering 22 activities, 17 participants, and 3 environments. 
\textbf{XRF55} is a large-scale and complex dataset for Wi-Fi-based human activity recognition. 
It contains 128,700 samples covering 55 human daily activities, 31 participants, and 4 environments. 
In our experiments, we retain all XRF55 activities and remove two overlapping activity classes from WiDAR 3.0, resulting in a total of 75 activity classes.

\subsection{Data Processing}

We follow the procedure described in the activity feature extractor to preprocess the CSI measurements and extract the corresponding signal representations using MATLAB.
To enable batch processing and ensure consistent model inputs, all samples are resampled to a fixed temporal length of 1,000 Wi-Fi packets before being fed into the activity feature encoder.
After preprocessing and flattening, the amplitude and phase representations have dimensions of $1000 \times 90$, whereas the DFS representation has dimensions of $1000 \times 121 \times 1$.

\subsection{Model and Environment Settings}

As described in Section~\ref{Methodology}, each signal encoder comprises two ViT-style Transformer layers, whereas the signal embedding extractor contains four self-attention blocks. 
The signal embedding aggregator and text embedding aggregator each consist of a single Transformer layer. 
The signal embedding aggregator includes an additional linear projection layer.
Both the signal and text modules produce 512-dimensional output embeddings.
For the LLM-based activity describer, we employ ChatGPT 5.5 as the backbone model.
The visual aids are provided with the original datasets and are used only as class-level references for generating activity descriptions.

During pretraining, the model is optimized for 60,000 iterations using Adam~\cite{kingma2014adam} with a learning rate of $10^{-4}$ and a batch size of 32. 
Training requires approximately 7 hours. 
The framework is implemented in PyTorch 2.0.2 with Python 3.9.18 and CUDA 11.8. 
All experiments are conducted on a Linux 24.04 platform equipped with an Intel(R) Core(TM) i9-13900 CPU, an NVIDIA GeForce RTX 4090 GPU with 24 GB of video memory.




\subsection{Zero-Shot Evaluation Protocol} 
We evaluate all zero-shot-compatible methods under a class-disjoint protocol. 
Specifically, in each trial, the activity categories are partitioned into a seen-class set and an unseen-class set, with no overlap between them. 
All model parameters are trained exclusively using Wi-Fi samples from the seen classes. 
Wi-Fi samples from the unseen classes are used only during final evaluation and are not involved in model training, hyperparameter selection, receiver selection, or checkpoint selection. 
Additionally, no classifier is trained or adapted using unseen-class Wi-Fi samples.
During inference, the descriptions of the unseen activities are introduced as fixed candidate semantic representations, and only to classify the held-out Wi-Fi samples.
Participants and environments are not treated as additional holdout factors in the current evaluation; therefore, the reported results specifically measure generalization across activity categories rather than cross-subject or cross-environment generalization.
We designate seven activity categories out of all 75 classes as unseen classes. 
The unseen categories are randomly selected in six independent trials. 

We use classification accuracy as the evaluation metric for the unseen-activity classification task, measuring the proportion of unseen activity samples that are correctly classified.
To reduce the influence of an exceptionally easy or difficult class partition, we report a one-sample-per-tail trimmed mean: for each method, the highest and lowest split accuracies are removed, and the remaining accuracies are averaged.

\subsection{Baselines}

We compare \methodname{} with five representative baselines on the WiFi-based HAR task. The baselines include \textbf{THAT}~\cite{li2021two}, a Transformer-based HAR system under supervised learning that achieves strong recognition performance. \textbf{OneFi}~\cite{xiao2021onefi}, a Transformer-based meta-learning framework designed to recognize previously unseen activities in a one-shot setting. \textbf{CLAR}~\cite{xiao2024diffusion}, a diffusion-based contrastive learning system for HAR. \textbf{FM-ZSL-IoT}~\cite{xue2024leveraging} and \textbf{Wi-CLIP}~\cite{zhang2025wi}, both employ Transformer-based contrastive learning for zero-shot WiFi-based HAR.

For all baselines, we implement the architectures and training procedures based on their original papers and publicly released code, when available. 
Necessary modifications are made to accommodate our datasets and zero-shot evaluation protocol.
For baselines that natively support signal-language zero-shot inference, i.e., FM-ZSL-IoT and Wi-CLIP, we retain their original semantic-inference procedures while adopting our activity splits and input data settings.
THAT, OneFi, and CLAR do not natively provide a mechanism for predicting unseen activity labels. They demonstrate that potentially transferable signal representations are insufficient for semantic zero-shot recognition when the system lacks a mechanism for associating unseen signal patterns with unseen activity labels.
Nevertheless, we pretrain their original architectures on the seen activities, use the resulting models as frozen signal encoders, and attach untrained, randomly initialized classifiers whose output dimensions correspond to the held-out activity classes. No Wi-Fi sample from unseen activities is used to optimize these classifiers.

\subsection{Results}

\begin{table}[!ht]
    \centering
    \begin{tabular}{cc}
      \toprule
        Method & Accuracy (\%) \\ \hline
        THAT~\cite{li2021two} & 16.60 \\
        OneFi~\cite{xiao2021onefi} & 11.33 \\
        CLAR~\cite{xiao2024diffusion} & 15.79 \\ 
        FM-ZSL-IoT~\cite{xue2024leveraging} & 23.86 \\ 
        Wi-CLIP~\cite{zhang2025wi} & 26.79 \\ \hline
        \textbf{\methodname{}} & \textbf{69.58} \\
      \bottomrule
    \end{tabular}
    \caption{Performance comparisons on unseen activities.}
    \label{comp_w_baselines}
\end{table}
\textbf{Comparisons with Baselines.}
Following the zero-shot evaluation protocol described above, all methods are evaluated on identical seen-unseen activity-class splits
The trimmed mean accuracy across these splits is reported in Table~\ref{comp_w_baselines}.
Among the methods that directly support zero-shot inference, \methodname{} achieves the best performance, with an accuracy of 69.58\%, substantially outperforming Wi-CLIP and FM-ZSL-IoT. 
The performance of THAT, OneFi, and CLAR under this setting reflects the absence of a learned correspondence between their extracted signal representations and the held-out activity labels.
\methodname{} achieves this performance by explicitly aligning Wi-Fi signal representations with natural-language activity descriptions.
Although FM-ZSL-IoT achieves an accuracy of 23.86\%, its fine-tuning stage relies on GAN-generated synthetic data, which may introduce distributional discrepancies and artifacts that limit further performance improvements~\cite{gong2025data}.
Wi-CLIP~\cite{zhang2025wi} achieves an accuracy of 26.79\%. 
However, its reliance on relatively uninformative single-sentence descriptions and a BERT-based text encoder that is not explicitly optimized for signal-language alignment~\cite{koroteev2021bert} limits its zero-shot inference capability.




\begin{figure}[!ht]
    \centering
    \includegraphics[width=0.85\columnwidth]{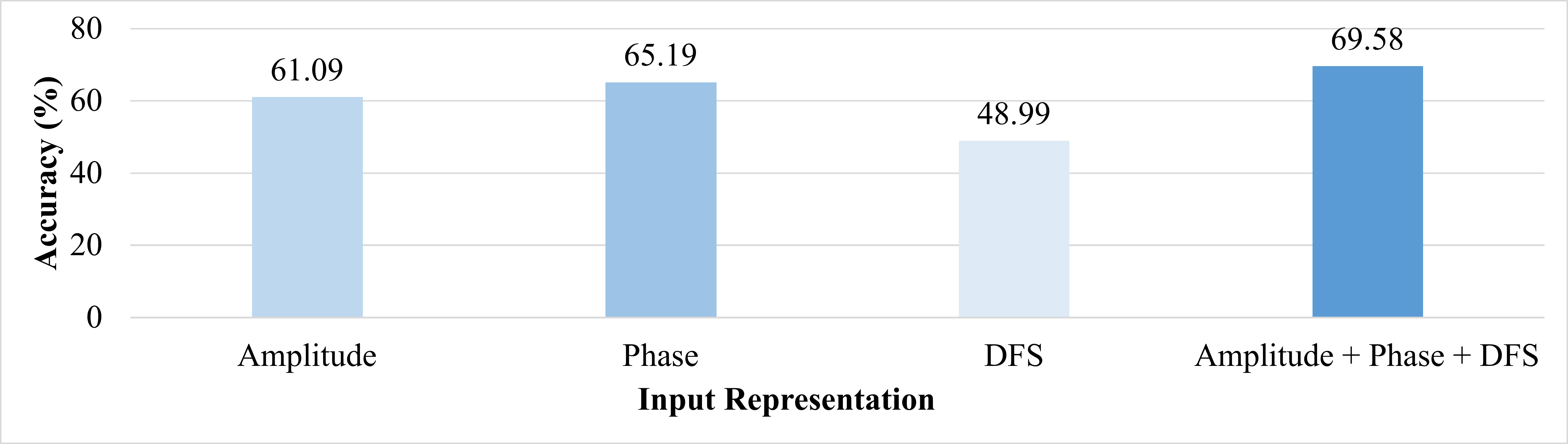}
    \caption{Ablation study: impact of signal representations.}
    \label{eff_of_signal_rep}
\end{figure}

\textbf{Ablation Studies.}
We first evaluate the contribution of individual CSI representations and their combinations through an ablation study. 
As shown in Figure~\ref{eff_of_signal_rep}, amplitude achieves an accuracy of 61.09\%, phase achieves the highest accuracy among the individual representations at 65.19\%, and DFS achieves an accuracy of 48.99\%. 
Combining all three representations yields the best accuracy of 69.58\%, improving upon the strongest individual representation by 4.39\%.
Amplitude and phase jointly capture the overall characteristics of human motion, while DFS contributes complementary frequency-domain information that further improves zero-shot recognition accuracy for unseen activities. 
Together, amplitude, phase, and DFS provide complementary motion information, yielding a more comprehensive representation that improves zero-shot recognition of unseen activities.

\begin{table}[!ht]
    \centering
    \begin{tabular}{cc}
      \toprule
        Text format & Accuracy (\%) \\ \hline
        Simple Activity Label & 48.49 \\
        Language Description & 69.58 \\ \hline
      \bottomrule
    \end{tabular}
    \caption{Ablation study: LLM-based activity describer.}
    \label{effect_of_act_des}
\end{table}


We further evaluate the effectiveness of the LLM-based activity describer. Specifically, we directly feed simple activity labels, such as ``walking'' and ``running,'' into the CLIP-based text encoder while keeping all other model components unchanged. As shown in Table~\ref{effect_of_act_des}, replacing the generated activity descriptions with simple labels leads to a substantial decrease in unseen-activity classification accuracy. This result indicates that the enriched descriptions provide more informative semantic supervision by explicitly characterizing activity-related motion attributes, thereby improving the model's ability to transfer to held-out activity classes.

\begin{table}[!ht]
    \centering
    \begin{tabular}{cc}
      \toprule
        Domain Discriminator & Accuracy (\%) \\ \hline
        Off & 63.68 \\
        On & 69.58 \\ \hline
      \bottomrule
    \end{tabular}
    \caption{Ablation study: domain discriminator.}
    \label{effect_of_domain_dis}
\end{table}
Moreover, we evaluate the contribution of the domain discriminator by removing it while keeping all other experimental settings and hyperparameters unchanged. 
As shown in Table~\ref{effect_of_domain_dis}, disabling the domain discriminator reduces the recognition accuracy from 69.58\% to 63.68\%. 
This result demonstrates that the domain discriminator improves unseen-activity recognition, suggesting that reducing dataset and environment-specific variations benefits the learned activity embeddings.

\section{Conclusion}

In this work, we propose \methodname{}, a contrastive signal-language alignment framework for zero-shot Wi-Fi-based human activity recognition. 
The framework leverages semantically rich textual representations and aligns them with CSI-derived signal embeddings in a shared latent space, thereby enabling the recognition of previously unseen activity categories without requiring labeled samples from those classes. 
Extensive experiments demonstrate that \methodname{} achieves strong zero-shot recognition performance and exhibits robust generalization to unseen activities.

\newpage

\bibliography{reference}

\newpage

\end{document}